\documentclass{article}
\usepackage{preprint,times}

\usepackage{amsmath,amsfonts,bm}

\def\eqref#1{equation~\ref{#1}}

\def\1{\bm{1}}

\DeclareMathAlphabet{\mathsfit}{\encodingdefault}{\sfdefault}{m}{sl}
\SetMathAlphabet{\mathsfit}{bold}{\encodingdefault}{\sfdefault}{bx}{n}

\usepackage{amsmath,amssymb}
\usepackage{graphicx}
\usepackage{booktabs}
\usepackage{array}
\usepackage{xcolor}
\usepackage{hyperref}
\usepackage{url}
\usepackage{microtype}

\title{GLIMPSE: Global--Local Indexing with Multi-Precision
Spectral Encoding for Explorable Video World Models}

\author{%
  \normalfont\begin{minipage}[t]{0.97\textwidth}
  \centering
  Zhufeng Xu$^{1,2}$\quad Henglei Lv$^{1,2}$\quad Bailin Deng$^{3}$\quad Yikang Ding$^{4}$\\[3pt]
  Xiaoqiang Liu$^{4}$\quad Guoxin Zhang$^{4}$\quad Hongbo Fu$^{5}$\quad Lin Gao$^{1,2,*}$\\[6pt]
  {\small
  $^{1}$Institute of Computing Technology, Chinese Academy of Sciences\\
  $^{2}$University of Chinese Academy of Sciences\quad $^{3}$Cardiff University\\
  $^{4}$Kling AI\quad $^{*}$Corresponding author: Lin Gao\\
  $^{5}$The Hong Kong University of Science and Technology\quad\href{https://rabberk.github.io/glimpse-project/}{Project page}}
  \end{minipage}%
}

\hypersetup{pdfauthor={Zhufeng Xu, Henglei Lv, Bailin Deng, Yikang Ding, Xiaoqiang Liu, Guoxin Zhang, Hongbo Fu, Lin Gao},
  pdftitle={GLIMPSE: Global-Local Indexing with Multi-Precision Spectral Encoding for Explorable Video World Models}}

\iclrfinalcopy
\begin{document}
\maketitle
\fancyhead{}

\begin{figure}[!ht]
  \centering
  \includegraphics[width=\linewidth]{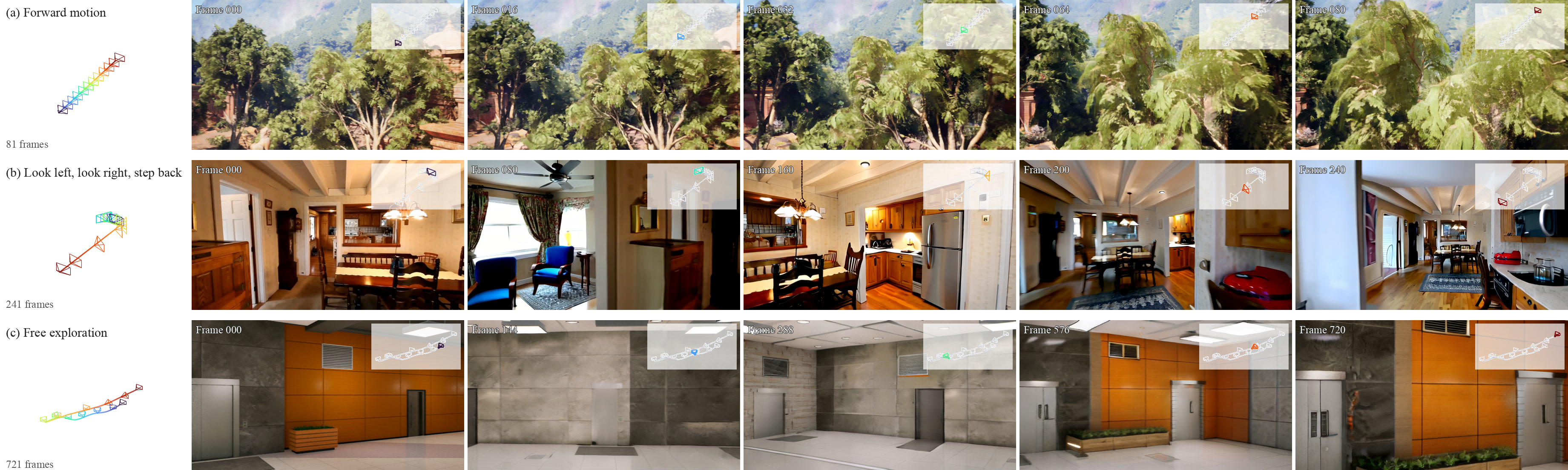}
  \caption{\textbf{Camera-controlled scene exploration with GLIMPSE.}
  From top to bottom: forward motion, left/right observation followed by
  backward motion, and free exploration over 81, 241, and 721 frames.
  Each row shows the requested camera trajectory on the left and five
  generated frames in temporal order. Insets highlight the corresponding
  camera pose, and colors indicate progression along the trajectory.}
  \label{fig:teaser}
\end{figure}

\begin{abstract}
A camera-controlled video world model must keep a scene consistent when the
camera returns to it. Memories that render stored features into the target
frame can turn depth and pose errors into ghosting or duplicated structure.
Positional transport avoids
that resampling by relocating a historical patch's address rather than its
content, but assigns full positional precision even to inaccurate projections.
We present GLIMPSE, a global--local memory interface with
band-wise multi-precision coordinates and geometry-conditioned spectral
attenuation to reduce sensitivity to inaccurate projected addresses.
Local patch memory supplies view-specific appearance, while recurrent global
memory provides scene context.
Footprint-guided retrieval selects historical patches under a fixed local read
budget. Under matched retrieval and read budgets, GLIMPSE improves geometric
correspondence and photometric consistency
over deterministic positional transport.
\end{abstract}

\begingroup
\raggedbottom
\section{Introduction}

An explorable video world model generates the next video window along
a requested camera trajectory. During extended exploration it must also remain
self-consistent: surfaces should retain their appearance when the camera
returns to them.
Revisiting previously observed regions requires retrieving historical appearance
and addressing it in the requested view. Yet geometric estimates are imperfect,
making the reliability of these positional addresses critical to consistent generation.

Prior work follows three main approaches. Frame- and view-retrieval memories retain selected historical
observations~\citep{contextasmemory,worldmem,vmem}. A limited set of views can
preserve scene identity but may omit details needed by a returning viewpoint.
Value-warping memories
instead reconstruct geometry and render stored features onto the target
grid~\citep{longtermspatialmemory,mirage}. This provides explicit geometric
alignment, but resolves estimated depth and pose into pixel positions, so
reprojection errors can propagate into generation as ghosting or duplicated
structure. Positional transport avoids this resampling by leaving each
historical patch exactly as it was observed and relocating only its address
within attention~\citep{ucm,mosaicmem}. Each token remains associated with its
original patch, without resampling historical features onto the target grid.

Positional transport preserves historical appearance, but relies on estimated
geometry to locate it in the target view. Errors in depth or camera pose can
misplace these positional addresses, with high-frequency rotary bands being
particularly sensitive to small coordinate errors. Standard positional transport
does not account for this varying geometric reliability. We therefore adapt the
precision and frequency content of each address to the reliability of its
underlying geometry, preserving fine positional detail when reliable and
reducing sensitivity to uncertain projections otherwise
(Figure~\ref{fig:stape-motivation}).

\begin{figure}[!t]
  \centering
  \includegraphics[width=\linewidth]{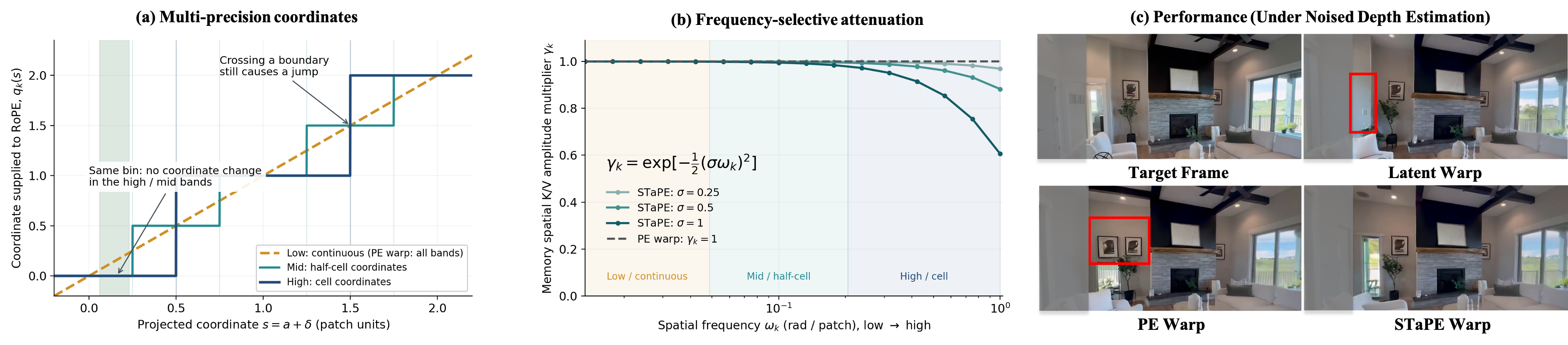}
  \caption{\textbf{Motivation for robust positional addressing.}
\emph{Left:} the coordinate supplied to RoPE against the continuous projected
coordinate, where the high and mid bands hold a cell and a half-cell value, so a
sub-cell error inside one bin changes nothing, while the low bands keep the
continuous offset. \emph{Middle:} the spectral multiplier $\gamma_k$ against
frequency, where PE warp keeps every band at full amplitude, whereas
Spatio-Temporal Aligned Positional Encoding (STaPE)
attenuates the highest frequencies first, more strongly as $\sigma$ grows.
\emph{Right}, in reading order: the reference view, latent value warping, PE
warping, and STaPE under the same four-patch address perturbation. Value
warping drops the doorway and PE warping duplicates wall art, whereas STaPE preserves
both.}
  \label{fig:stape-motivation}
\end{figure}
We introduce \textbf{GLIMPSE}, \textbf{G}lobal--\textbf{L}ocal
\textbf{I}ndexing with \textbf{M}ulti-\textbf{P}recision \textbf{S}pectral
\textbf{E}ncoding, a global--local memory framework for explorable video
generation (Figure~\ref{fig:teaser}). In our framework, a frozen recurrent memory provides
global context, while a local pointer memory retrieves original historical
patches through projected footprint coverage under a fixed per-frame read
budget. The two memories are accessed through separately gated readouts.
To address retrieved patches in the requested view, we introduce
Spatio-Temporal Aligned Positional Encoding (STaPE), which represents projected
locations at band-specific coordinate precisions and regulates their spectral
strength using geometric reliability cues. This design reduces sensitivity to
projection errors without resampling historical appearance.

Our contributions are threefold:
\begin{itemize}
  \item We propose a \textbf{geometry-addressed global--local memory interface}
  that separates scene-level recurrent context from view-specific appearance.
  Footprint-guided retrieval selects original historical patches under a fixed
  local read budget. Separate gated
  readouts combine the two memories without making global context compete
  with local evidence in the same attention normalization.
  \item We introduce \textbf{multi-precision positional addressing}, assigning
  cell, half-cell, and continuous coordinates to different rotary frequency
  bands. Fine-frequency addresses remain unchanged under perturbations within
  a bin, while continuous lower-frequency coordinates retain sub-cell
  placement without adding memory tokens.
  \item We develop \textbf{geometry-conditioned spectral attenuation}, using
  depth smoothness and temporal lag to regulate spatial frequency
  components of memory keys and values. Together with multi-precision
  addressing, it controls the precision and strength of transported addresses.
  Matched ablations establish their complementary benefits over deterministic
  positional transport in geometric correspondence and photometric consistency.
\end{itemize}
\par\newpage
\endgroup

\section{Related Work}

\subsection{Interactive Video World Models}

Interactive video world models generate observations in response to user
actions while keeping the environment coherent across interaction steps.
In camera-directed systems, the action specifies a trajectory and the
observation is the next video window. InSpatio-World develops this
autoregressive trajectory-control interface, while AlayaWorld, Wonder,
DreamX-World, and Infinite-World target long-running playable or
general-purpose environments
\citep{inspatioworld,alayaworld,wonder,dreamxworld,infiniteworld}.  GLIMPSE
focuses on retaining historical appearance and retrieving it for a new
camera query. Vid2World adapts pretrained video diffusion models to causal
interactive generation~\citep{vid2world}, and MiniWorld instead studies a
reproducible streaming training recipe from scratch~\citep{miniworld}.
GLIMPSE studies the memory interface within a pretrained video generator.

\subsection{Camera Control and Positional Alignment}

Camera-controllable generation conditions synthesis on an explicit viewpoint
trajectory. CameraCtrl injects camera-ray conditioning~\citep{cameractrl},
whereas TrajectoryCrafter jointly conditions on point-cloud renders and
source video through a dual-stream diffusion model~\citep{trajectorycrafter}.
GEN3C conditions generation on a
point-cloud cache rendered at the requested camera~\citep{gen3c}. UCPE combines Relative Ray
Encoding of 6-DoF pose, intrinsics, and lens distortion with Absolute
Orientation Encoding of pitch and roll through a lightweight spatial-attention
adapter~\citep{ucpe}. PRoPE expresses relative camera geometry as projective
positional encoding inside attention~\citep{prope}. CameraSquad combines
decoupled content and camera control with dual-mode cross-view attention
for consistent parallel multi-trajectory generation~\citep{camerasquad}.
These encode geometry to
\emph{control} generated target tokens.  UCM applies the same VAE and
patchification to historical and target latents, then warps positional encoding
in time, assigning target space-time coordinates to clean historical tokens
while leaving their content unchanged~\citep{ucm}.
Integrated positional encoding attenuates Fourier features according to
spatial uncertainty~\citep{mipnerf}, and our work instead considers
geometry-conditioned positional addressing for retrieved video memory.

\subsection{Persistent and Spatial Memory}

Recent-context methods improve how a finite frame history is packed or
weighted~\citep{framepack,historyguidance,lyra2}.  Persistent-memory methods
retain information beyond that window: Context as Memory and WorldMem retrieve
stored frames, VMem retrieves views through a surfel index of the queried
surface, and SPMem combines working, spatial, and episodic
stores~\citep{contextasmemory,worldmem,vmem,longtermspatialmemory}.  CUT3R
maintains a recurrent state for streaming 3D perception and MeMix updates that
state selectively to reduce forgetting~\citep{cut3r,memix}. GLIMPSE uses these
frozen components for its global branch. Long-context state-space video world models
combine blockwise state-space scanning with local attention~\citep{longcontextssm}.
GIM-World compresses history into learned geometry-aware implicit memory and
prunes observations~\citep{gimworld}. In contrast to this compressed history,
LPM retains the provenance of individual source patches. WorldPlay reconstructs
context from past frames and temporally reframes old observations to maintain
geometric consistency~\citep{worldplay}.

A further line of work makes spatial memory directly
addressable.  Mirage persists latent features at world points and renders their
values onto a target grid, MosaicMem indexes localized patches and aligns them
through warped RoPE or warped latents, and SPMem renders explicit spatial-memory
values in the target view alongside sparse episodic
frames~\citep{mirage,mosaicmem,longtermspatialmemory}.  The value-warping variants commit appearance to a target raster before
attention, whereas positional-warping variants preserve separate source
features. AnchorWeave retrieves local geometric memories by trajectory coverage and
renders them as anchor conditions~\citep{anchorweave}, and WorldTrace keeps a
compressed cache addressable by assigning summary slots virtual
positions~\citep{worldtrace}. Concurrently, ReWorld bounds long-horizon memory
with a pose-indexed landmark bank under a fixed KV budget~\citep{reworld}.
LPM uses the positional-warping route with a fixed per-target-frame token
budget: footprint coverage selects views, and a separate allocator gathers
the original patches that enter attention.

\section{Method}

\begin{figure}[t]
  \centering
  \includegraphics[width=\linewidth]{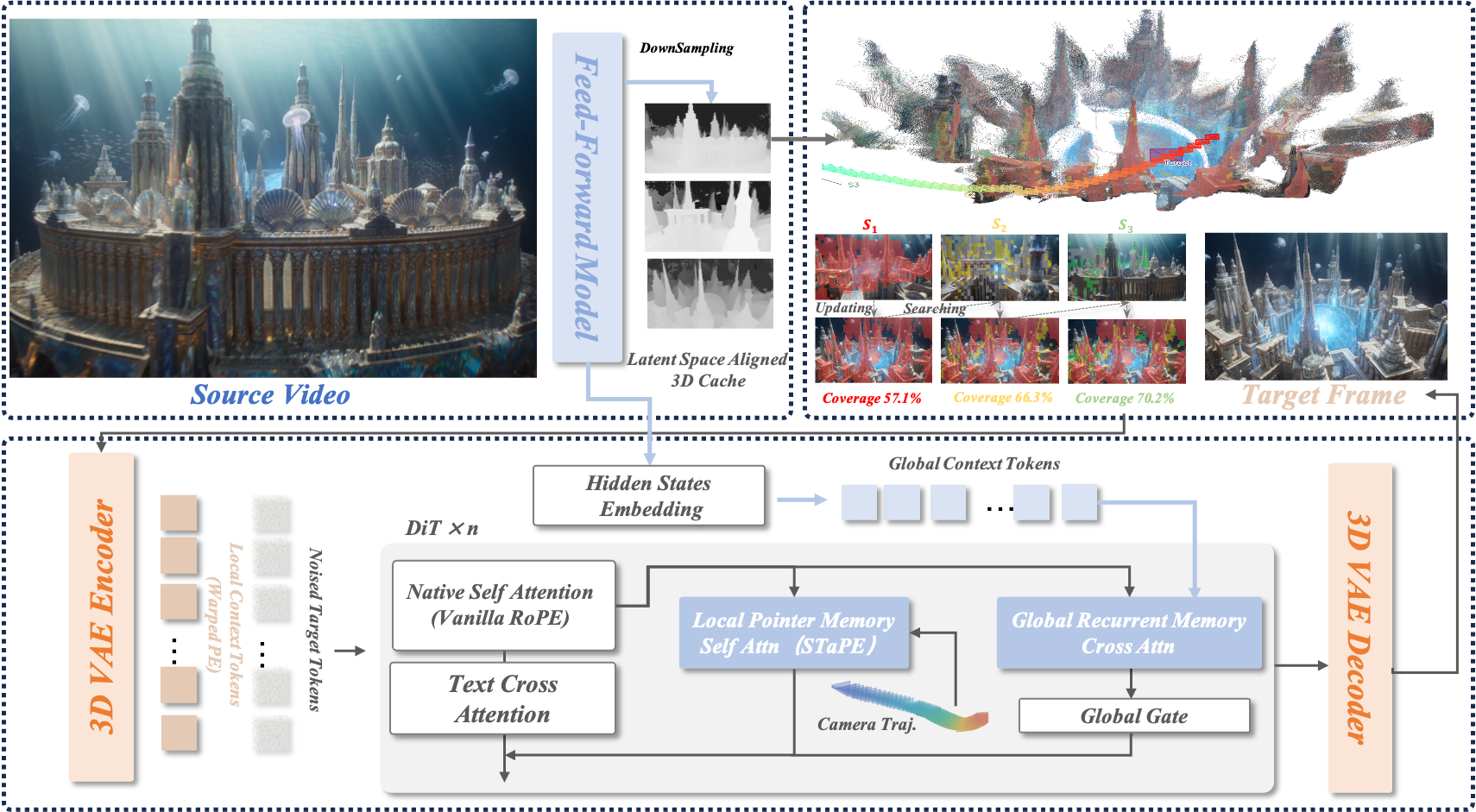}
  \caption{Overview of GLIMPSE. Geometry-guided greedy retrieval selects local
  context from the source history. STaPE aligns the positions of retrieved
  latent tokens with the requested target view, while recurrent global
  context provides complementary scene memory to the video generation model.}
  \label{fig:method-overview}
\end{figure}

\subsection{Framework Overview}

Figure~\ref{fig:method-overview} illustrates the overall framework.
At exploration step $t$ the user specifies a camera action $a_t=(T_t^q,K_t^q)$:
the requested target pose and intrinsics. Let $x_t^{\mathrm{src}}$ denote the
current source observation and $S_t=(G_t,L_t)$ the global and local memories.
The local memory also stores camera and time indices. Generation and memory
updates follow
\begin{equation}
 a_t=(T_t^q,K_t^q),\qquad
 o_{t+1}\sim p_\theta\!\left(o\mid x_t^{\mathrm{src}},a_t,G_t,L_t\right),
 \qquad
 S_{t+1}=\mathcal U(S_t,o_{t+1},a_t).
 \label{eq:world-state-interface}
\end{equation}
Each update advances GRM and appends observations and geometry to LPM.
Memory read length is fixed, but storage, indexing, and geometry estimation
grow with history. Appendix~\ref{sec:retrieval-scaling} profiles geometry,
pointer-cache, and denoising costs across horizons.

\subsection{Global--Local Memory}

The GRM $G_t$ is a frozen CUT3R~\citep{cut3r} recurrent state built by
processing selected observations in chronological order, using the
parameter-free MeMix update to combine retained state tokens with each new observation. It provides a fixed-size summary of the scene and its history.
The LPM is a geometry-indexed interface over observation history,
\begin{equation}
 I_t=\{(\mathbf X_j,\mathbf d_j,\iota_j,c_j,\alpha_j)\}_{j\in\mathcal H_t},
 \quad R_{t,f}^q=\mathcal Q(I_t;T_{q,f},K_{q,f}),
 \quad |R_{t,f}^q|\leq M_f,
 \label{eq:lpm-state}
\end{equation}
where $\mathbf X_j$ and $\mathbf d_j$ denote the world position and viewing ray,
$\iota_j$ identifies the original observation and latent patch, and $c_j$ and
$\alpha_j$ record confidence and temporal lag. We rebuild $I_t$ from the available
history geometry at each query and limit each target frame's read to one latent
frame's worth of tokens. The DiT receives the original patch tokens selected by
$R_{t,f}^q$, not the index itself. We rank candidate views by projected footprint
coverage, then select original patch tokens from these views based on their
projected centers.

LPM tokens have explicit target-view coordinates, and GRM tokens are addressed by content. We read GRM through a separately gated cross-attention, so global tokens do not compete with local evidence for attention weight.

\paragraph{Retrieval-to-read procedure.}
Let $H_p\times W_p$ be the shared patch-embedding grid and $M=H_pW_p$. For each
target frame $q$, candidate views are jointly ranked by camera proximity and
temporal lag, prioritizing source observations over generated history. Using the
occupancy masks $F_i$ of projected patch quadrilaterals, greedy selection adds
the view covering the most uncovered cells at each round. In selected-view order the
allocator keeps the first positive-depth, in-bounds projected center for each
target cell and fills the remaining capacity up to $M$ with unused original
tokens. These fillers may be redundant or outside the target frustum.
The allocator preserves individual token features and applies no target-depth
test. Table~\ref{tab:visibility-planned}
compares this allocator with center-hit and causal z-buffer controls under
a matched token budget.

At each denoising step, history and main-stream latents share the DiT patch
embedding. Each backbone layer updates pointer tokens at timestep zero using
source-coordinate native RoPE and a $\tanh(g_\ell)$ gate, then processes the
main stream. Target queries read the associated pointer K/V through STaPE and
GRM through a separate softmax. The two outputs form the conditioning residual
(Equation~\ref{eq:adapter-residual}).
Appendix~\ref{sec:retrieval-details} provides the implementation settings
and retrieval details.

\subsection{Spatio-Temporal Aligned Positional Encoding}
\label{sec:stape}

STaPE places a historical patch in the continuous spatial
coordinate system of its associated target frame.  For latent-cell center $\mathbf x_h$, historical
depth $D_h$, historical pose $T_h$, and target camera $(T_q,K_q)$,
\begin{equation}
 (u_h^q,v_h^q)=\operatorname{project}(K_q,K_h,T_q,T_h,D_h,\mathbf x_h),
 \qquad \phi(h)=q.
 \label{eq:stape-coordinate}
\end{equation}
Here $T_h,T_q$ are camera-to-world poses in the same DA3 scene gauge, and
$K_h,K_q$ are intrinsics at the coordinate resolution used in the projection.
For patch-grid center $\mathbf x_h=(c+\tfrac12,r+\tfrac12,1)^\top$,
the implementation unprojects
$\mathbf X_h=R_h(d_hK_h^{-1}\mathbf x_h)+\mathbf t_h$ and transforms it to
the target as $\mathbf X_q=R_q^\top(\mathbf X_h-\mathbf t_q)$.
Homogeneous division of $K_q\mathbf X_q$ gives image coordinates, which are
scaled to target patch-grid units and shifted by $-\tfrac12$ so integer
coordinates denote patch centers.
Appendix~\ref{sec:projection-refresh} specifies depth resizing and the
separate handling of projected support and budget fillers.
The projection retains a continuous center estimate, and the band-dependent
coordinate rule below then selects its encoding precision. Membership
$\phi(h)=q$ supplies temporal alignment through the frame-local adapter mask.
No separate temporal rotary phase is introduced in this adapter.

Target and retrieved history latents pass through the same DiT patch embedding
$E$, adapter normalization and projection $A$, and key/value projections.
The pointer refresh reuses native backbone weights, whereas each layer's
adapter has separate normalization and projections initialized from the
corresponding DiT block.
Appendix~\ref{sec:projection-refresh} gives the execution order and the
detailed shadow update and residual equations.
Let $\mathbf h_i^0=E(\mathbf z_i)$ and let $\mathbf h_i^\ell$ denote
the hidden feature supplied to the local read at layer $\ell$ after the
above processing. The content tensors before positional encoding are
\begin{equation}
 \begin{alignedat}{2}
 \mathbf f_i&=A(\operatorname{Norm}(\mathbf h_i^\ell)),
 &\qquad Q_q^{\mathrm{pre}}&=W_Q\operatorname{Norm}_Q(\mathbf f_q),\\
 K_i^{\mathrm{pre}}&=W_K\operatorname{Norm}_K(\mathbf f_i),
 &\qquad V_i^{\mathrm{pre}}&=W_V\mathbf f_i.
 \end{alignedat}
 \label{eq:stape-key-value}
\end{equation}
Here $\operatorname{Norm}$, $\operatorname{Norm}_Q$, and
$\operatorname{Norm}_K$ are the adapter's learned LayerNorms, applied before
head reshaping. 
Let $P_q$ and $P_h$ be the target and historical token's source-camera projective
matrices, incorporating intrinsics in a shared scene gauge. PRoPE partitions each head into a
$D/2$ camera block and two $D/4$ spatial blocks~\citep{prope}.
Appendix~\ref{sec:projection-refresh} defines the corresponding Q, K/V, and O
basis matrices. Within either spatial block, pair $k$ consists of channels $(k,k+m)$,
$m=D/8$, and
\begin{equation}
 R_k(a)=
 \begin{bmatrix}
 \cos(\omega_k a)&\sin(\omega_k a)\\
 -\sin(\omega_k a)&\cos(\omega_k a)
 \end{bmatrix},
 \qquad \omega_k=100^{-k/m}.
 \label{eq:stape-pair}
\end{equation}
The adapter and native DiT backbone use their respective frequency schedules.

\paragraph{Geometry-conditioned spectral regularization.}
We attenuate both channels of each spatial rotary pair by
$\gamma_k=\exp[-(\sigma\omega_k)^2/2]$, motivated by Gaussian perturbations
of continuous coordinates (Appendix~\ref{sec:spectral-scope}). Collecting these
factors in $\Gamma(\sigma)$ gives the operator
\begin{equation}
 \begin{aligned}
 D_\sigma&=\operatorname{diag}\!\left(I_{D/2},\Gamma(\sigma),\Gamma(\sigma)\right),\\
 \mathcal P_{h,\sigma}^{KV}(u,v)&=D_\sigma B_h^{KV}(u,v).
 \end{aligned}
 \label{eq:stape-operator}
\end{equation}
Memory K/V use $P_h^{-1}$ with target-projected spatial coordinates,
while attenuation leaves camera components and target queries unchanged.
Let $a_q=(u_q,v_q)$ and let $Q_q$ denote the target query transformed
by the standard PRoPE operator $B_q^Q(a_q)$.
For local memory $H_q=\{h:\phi(h)=q\}$, the positional transformation and
memory attention are defined as follows:
\begin{equation}
 \begin{aligned}
 (K_h^{\mathrm{pos}},V_h^{\mathrm{pos}})
 &=\mathcal P_{h,\sigma_h}^{KV}(u_h^q,v_h^q)(K_h^{\mathrm{pre}},V_h^{\mathrm{pre}}),\\
 Y_q&=B_q^O(a_q)\operatorname{Attn}
 \left(Q_q,K_{H_q}^{\mathrm{pos}},V_{H_q}^{\mathrm{pos}}\right).
 \end{aligned}
 \label{eq:stape-attention}
\end{equation}
Superscripts $\mathrm{pre}$ and $\mathrm{pos}$ denote features before and
after positional transformation. We determine each token's attenuation width
from local depth smoothness and temporal lag:
\begin{equation}
 \begin{aligned}
 c_h&=\operatorname{clip}_{[0,1]}
 \exp\!\left[-4\,
 \frac{|\Delta_x d_h|+|\Delta_y d_h|}{\max(d_h,10^{-3})}\right],\\
 \alpha_h&=1-\bar t_h,\\
 \sigma_h&=\lambda_c[1-c_h]_++\lambda_a[\alpha_h]_+ .
 \end{aligned}
 \label{eq:stape-attenuation}
\end{equation}
Here $\bar t_h\in[0,1]$ is the normalized observation time. Larger widths
suppress high spatial frequencies without changing the projected coordinate.
At $\sigma_h=0$, attenuation is disabled, while the multi-precision coordinate
rule remains active.

\paragraph{Multi-precision coordinates.}
For pair index $k$, let $b_1=\lceil m/3\rceil$ and
$b_2=\lceil2m/3\rceil$. We replace the coordinate in each spatial rotation by
\begin{equation}
 q_k(a)=
 \begin{cases}
 \operatorname{round}_{\mathrm{even}}(a),&k<b_1,\\
 \tfrac12\operatorname{round}_{\mathrm{even}}(2a),&b_1\leq k<b_2,\\
 a,&k\geq b_2 .
 \end{cases}
 \label{eq:stape-footprint}
\end{equation}
Here $\operatorname{round}_{\mathrm{even}}$ rounds to the nearest integer,
ties to even. High-frequency pairs
use cell coordinates, middle pairs use half-cell coordinates, and low-frequency
pairs retain continuous offsets. In the quantized bands, spatial phases remain
unchanged within each bin and change discontinuously across bin boundaries.
For this variant, $R_k(a)$ in Equation~\ref{eq:stape-pair} is replaced by
$R_k(q_k(a))$ and the attenuation in Equation~\ref{eq:stape-operator}
is unchanged.
Source-token identity is preserved throughout: the operator transforms feature
channels without spatially resampling or averaging appearance between tokens.

\subsection{Training and Data Construction}

We train on a mixture of SpatialVID~\citep{spatialvid},
RealCam-Vid~\citep{realcamvid}, ReCamMaster~\citep{recammaster}, and
Context-as-Memory~\citep{contextasmemory}. Cached DA3~\citep{da3} pose, intrinsics, and depth are the
only geometry source, and a cache miss skips the sample instead of invoking online
estimation. Three augmentations stress the memory interface.
\emph{Future-as-memory} uses real post-target observations as training-only LPM
donors with relabeled memory times. Inference uses only source observations and
completed generated windows. \emph{Synthetic revisitation} builds
complementary old-view memory from multiple keyframes under a shared camera
perturbation, and \emph{fast motion} temporally subsamples the target at stride
two or three while keeping the dense history. Sampling weights, offsets,
probabilities, and eligibility rules are specified in
Appendix~\ref{sec:training-details}. Table~\ref{tab:augmentation-ablation}
isolates the augmentations through separately trained variants, including
causal-only donors, evaluated under the same causal inference protocol.

Training initializes the dense DiT from Wan2.2-TI2V-5B, runs at
$256\!\times\!448$, and continues at $352\!\times\!640$. Only the target latent
is noised. We parameterize the interpolation by the noise level
$s=\sigma_j$, selected from the scheduler,
and $t_j=1000s$ is the timestep supplied to the DiT. For
$\epsilon\sim\mathcal N(0,I)$,
\begin{equation}
 \begin{aligned}
 Z_s&=(1-s)Z_0+s\epsilon,\\
 \mathcal L_{\mathrm{FM}}&=
 \mathbb E_{j,\epsilon}\!\left[
 w_j\left\|v_\theta(Z_s,t_j;x^{\mathrm{src}},a,G,L)
 -(\epsilon-Z_0)\right\|_2^2\right].
 \end{aligned}
 \label{eq:full-flow-matching}
\end{equation}
Thus $v_\theta$ predicts $\partial Z_s/\partial s=\epsilon-Z_0$.
Inference integrates in $s$ using the Euler update
$Z_{s'}=Z_s+(s'-s)v_\theta(Z_s,1000s;\cdot)$, ending at $s'=0$.
Appendix~\ref{sec:flow-schedule} gives the shifted nodes and loss weights.

After each window, DA3 re-estimates source and generated-history geometry with
the source anchor fixed. Subsequent commands, intrinsics, and camera history
are rebased onto this estimated trajectory.

\section{Experiments}
\subsection{Experimental Setup}
GLIMPSE uses Wan2.2-TI2V-5B~\citep{wan}, cached DA3 geometry~\citep{da3},
and frozen CUT3R recurrent context~\citep{cut3r,memix}.
Camera control uses trajectory errors, CLIP-T/F, FID, and FVD.
WorldScore~\citep{worldscore} evaluates quality and consistency, while Y-traj.
also measures revisit correspondences and RGB agreement.
We compare against UCM~\citep{ucm}, SPMem~\citep{longtermspatialmemory},
GEN3C~\citep{gen3c}, VMem~\citep{vmem}, and the concurrent
ReWorld~\citep{reworld}, following their public inference recipes. These
comparisons evaluate complete systems, while matched internal ablations
isolate the contribution of STaPE.
SPMem denotes the spatial-memory video world model of \citet{longtermspatialmemory}.
Training settings, metric definitions, and comparison details are collected
in Appendix~\ref{sec:experiment-protocol}. 

\newcommand{\cameraWorldScoreCaption}{WorldScore components for the camera-control experiment.
Scores are normalized to 0--100 and higher is better, including transformed
error scores. The mean weights all nine components equally. Bold marks the
best score.}
\newcommand{\cameraWorldScoreBody}{%
\setlength{\tabcolsep}{0pt}
\begin{tabular}{@{}>{\raggedright\arraybackslash}p{0.28\linewidth}*{6}{>{\raggedleft\arraybackslash}p{0.12\linewidth}}@{}}
\toprule
Metric & \makebox[\linewidth][r]{GLIMPSE (ours)} & UCM & SPMem & GEN3C & VMem & ReWorld \\
\midrule
Camera & \textbf{91.032} & 90.245 & 81.273 & 82.510 & 81.294 & 84.740 \\
Text & \textbf{88.983} & 87.579 & 87.509 & 83.072 & 71.927 & 88.703 \\
3D reprojection & 83.930 & 77.286 & 64.313 & 75.779 & 50.567 & \textbf{85.149} \\
Flow AEPE & \textbf{89.752} & 87.874 & 79.693 & 84.519 & 64.088 & 85.933 \\
CLIP-IQA+ & \textbf{35.958} & 33.542 & 22.470 & 28.228 & 17.109 & 34.098 \\
Aesthetic quality & \textbf{15.357} & 13.081 & 11.010 & 13.875 & 11.920 & 13.882 \\
Motion magnitude & 34.026 & 33.702 & 33.805 & \textbf{37.658} & 28.592 & 34.132 \\
Motion smoothness & \textbf{66.605} & 62.055 & 65.267 & 61.899 & 44.810 & 66.227 \\
Style consistency & 94.380 & \textbf{94.488} & 89.305 & 93.297 & 85.011 & 92.012  \\
\midrule
Mean (9 metrics) & \textbf{66.669} & 64.428 & 59.405 & 62.315 & 50.591 & 64.986 \\
\bottomrule
\end{tabular}}

\subsection{Camera-Control Evaluation}

\begin{figure}[t]
\centering
\includegraphics[width=\linewidth]{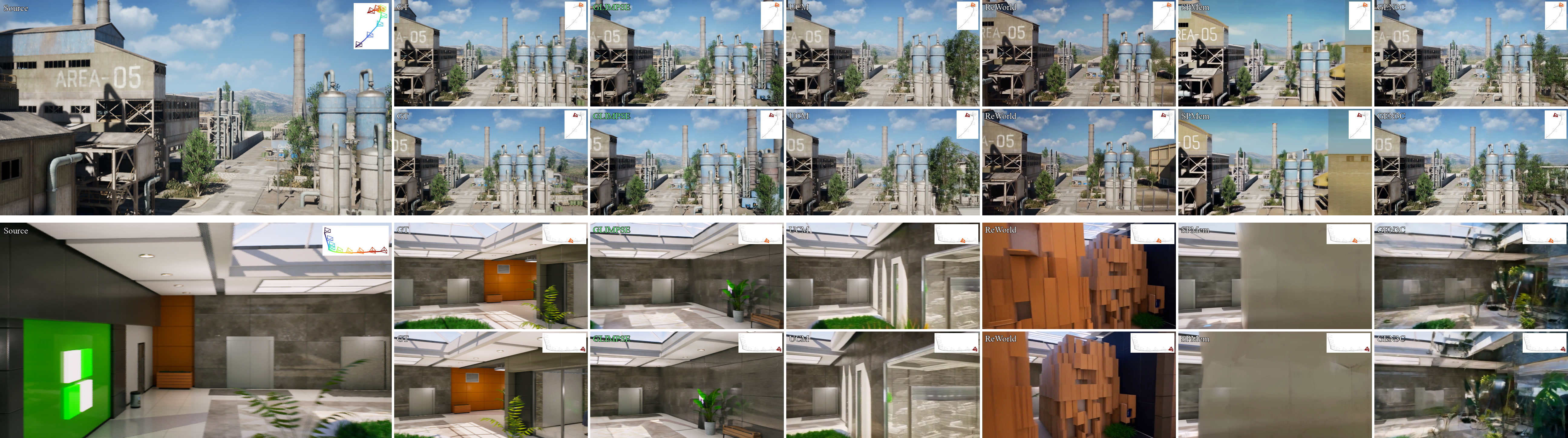}
\caption{Camera-trajectory replay on two scenes. The enlarged source image
is followed by GT, GLIMPSE, UCM, ReWorld, SPMem, and GEN3C (left to right), with two selected
time steps per scene. Insets show the requested camera trajectory and current
pose.}
\label{fig:camera-replay-qualitative}
\end{figure}

Table~\ref{tab:camera-control} summarizes camera diagnostics and
 video-quality metrics~\citep{recammaster}. GLIMPSE attains the lowest rotation errors and the lowest translation-direction error, together with the best
CLIP-T, CLIP-F and FID. UCM achieves lower translation NRMSE, while ReWorld
achieves lower FVD.
Table~\ref{tab:worldscore-subset} reports the WorldScore components on the
same subset: GLIMPSE leads six of the nine and the nine-component mean,
ReWorld ranks second overall and leads on 3D reprojection, while UCM leads on
style consistency. Figure~\ref{fig:camera-replay-qualitative} shows qualitative replay comparisons.

\begin{table}[t]
\centering
\begin{minipage}[t]{0.49\linewidth}
\vspace{0pt}
\caption{Camera-control and video-quality metrics. Rotation and direction
errors are in degrees. Bold marks the best score.}
\label{tab:camera-control}
\end{minipage}\hfill
\begin{minipage}[t]{0.49\linewidth}
\vspace{0pt}
\caption{\cameraWorldScoreCaption}
\label{tab:worldscore-subset}
\end{minipage}
\par\smallskip
\begin{minipage}[t]{0.49\linewidth}
\vspace{0pt}
\centering\tiny
\setlength{\tabcolsep}{0.8pt}
\renewcommand{\arraystretch}{1.222222}
\begin{tabular*}{\linewidth}{@{\extracolsep{\fill}}lrrrrrr@{}}
\toprule
Metric & GLIMPSE (ours) & UCM & SPMem & GEN3C & VMem & ReWorld \\
\midrule
Mean rot.$\downarrow$ & \textbf{0.77} & 0.88 & 5.56 & 1.29 & 7.87 & 2.94 \\
Endpoint rot.$\downarrow$ & \textbf{1.28} & 1.48 & 5.68 & 2.95 & 9.50 & 4.78 \\
Trans. NRMSE$\downarrow$ & 0.102 & \textbf{0.059} & 0.318 & 0.342 & 0.525 & 0.424 \\
Trans. dir.$\downarrow$ & \textbf{2.14} & 5.54 & 9.20 & 9.31 & 12.42 & 8.03 \\
CLIP-T$\uparrow$ & \textbf{30.445} & 29.486 & 28.641 & 29.041 & 26.357 & 30.012 \\
CLIP-F$\uparrow$ & \textbf{99.123} & 99.006 & 98.808 & 98.912 & 97.509 & 99.101 \\
FID$\downarrow$ & \textbf{62.218} & 62.387 & 65.487 & 63.874 & 89.887 & 62.307 \\
FVD$\downarrow$ & 863.838 & 877.592 & 940.247 & 903.112 & 1044.543 & \textbf{854.19} \\
\bottomrule
\end{tabular*}
\end{minipage}\hfill
\begin{minipage}[t]{0.49\linewidth}
\vspace{0pt}
\centering\tiny
\cameraWorldScoreBody
\end{minipage}
\end{table}

\subsection{Y-traj. Evaluation}
We introduce the \emph{Y-traj.} protocol to test whether a model can combine
complementary observations in memory. Each sequence observes the
left and right sides of a scene, then returns to the initial heading while
moving backward to a wider view, so the final frame can contain regions seen
at different times that no single earlier view covers. A trajectory that
merely retraces its own path never poses that problem.
Table~\ref{tab:worldscore-overview} reports
WorldScore results on 100 such cases. GLIMPSE attains the best value on seven
of the ten components and the highest ten-component mean. The remaining three
go to different systems: UCM on text alignment, ReWorld on DINO consistency,
and SPMem on motion smoothness.

\begin{table}[t]
\centering
\begin{minipage}[t]{0.49\linewidth}
\vspace{0pt}
\caption{WorldScore components on 100 Y-traj. cases. Scores are normalized to
0--100 and higher is better, including transformed errors. The mean weights
all ten components equally.}
\label{tab:worldscore-overview}
\end{minipage}\hfill
\begin{minipage}[t]{0.49\linewidth}
\vspace{0pt}
\caption{Revisit correspondences on 100 Y-traj. cases. @1/@5 use the last
one/five pairs per branch. Counts are in thousands (K) per pair, and RGB MAE uses
$[0,1]$ colors.}
\label{tab:ytraj-matching}
\end{minipage}
\par\smallskip
\begin{minipage}[t]{0.49\linewidth}
\vspace{0pt}
\centering\tiny
\setlength{\tabcolsep}{0pt}
\begin{tabular}{@{}>{\raggedright\arraybackslash}p{0.28\linewidth}*{6}{>{\raggedleft\arraybackslash}p{0.12\linewidth}}@{}}
\toprule
Metric & \makebox[\linewidth][r]{GLIMPSE (ours)} & UCM & SPMem & GEN3C & VMem & ReWorld \\
\midrule
CLIP-IQA+ & \textbf{21.826} & 14.161 & 11.075 & 14.631 & 12.113 & 20.922 \\
Aesthetic quality & \textbf{32.323} & 24.179 & 18.597 & 16.857 & 17.905 & 31.493 \\
Text alignment & 81.703 & \textbf{82.122} & 81.455 & 80.903 & 78.199 & 80.019 \\
Style consistency & \textbf{88.531} & 80.825 & 72.095 & 85.679 & 72.838 & 79.306\\
DINO consistency & 83.390 & 80.912 & 77.655 & 80.453 & 73.990 & \textbf{84.307} \\
CLIP consistency & \textbf{94.062} & 93.842 & 93.105 & 93.367 & 92.231 & 93.065 \\
Motion magnitude & \textbf{67.417} & 66.050 & 57.270 & 59.862 & 52.451 & 64.114 \\
Flow AEPE & \textbf{74.984} & 74.644 & 67.875 & 73.798 &  63.090& 74.578 \\
Motion smoothness & 33.685 & 30.847 & \textbf{36.094} & 30.357 & 30.917 & 30.281 \\
Camera control & \textbf{87.596} & 86.573 & 80.739 & 83.268 & 74.142 & 81.855 \\
\midrule
Ten-component mean & \textbf{66.552} & 63.415 & 59.596 & 61.918 & 56.788 & 63.994 \\
\bottomrule
\end{tabular}
\end{minipage}\hfill
\begin{minipage}[t]{0.49\linewidth}
\vspace{0pt}
\centering\tiny
\setlength{\tabcolsep}{0.8pt}
\renewcommand{\arraystretch}{1.333333}
\begin{tabular*}{\linewidth}{@{\extracolsep{\fill}}lrrrrrr@{}}
\toprule
Metric & GLIMPSE (ours) & UCM & SPMem & GEN3C & VMem & ReWorld \\
\midrule
Mat. Pix.@1 (K) $\uparrow$ & \textbf{5.602} & 5.307 & 2.592 & 4.223 & 2.138 & 4.799 \\
Mat. Pix.@5 (K) $\uparrow$ & \textbf{5.898} & 5.335 & 2.621 & 4.298 & 2.733 & 4.296 \\
Inliers@1 (K) $\uparrow$ & \textbf{4.846} & 4.338 & 2.149 & 3.722 & 1.430 & 3.704 \\
Inliers@5 (K) $\uparrow$ & \textbf{5.067} & 4.373 & 2.156 & 3.791 &  1.826& 3.717 \\
Inlier rate@1 (\%) $\uparrow$ & \textbf{82.301} & 75.752 & 80.048 & 77.810 & 72.885 & 72.991 \\
Inlier rate@5 (\%) $\uparrow$ & \textbf{79.353} & 78.922 & 76.440 & 77.133 & 72.813 & 72.744 \\
RGB MAE@1 $\downarrow$ & 0.093 & 0.095 & 0.102 & \textbf{0.079} & 0.147 & 0.099 \\
RGB MAE@5 $\downarrow$ & 0.091 & 0.097 & 0.102 & \textbf{0.086} & 0.152 & 0.099 \\
\bottomrule
\end{tabular*}
\end{minipage}
\end{table}

\paragraph{Revisit correspondences.}
We measure cross-view consistency by matching the earlier left/right observations
to the final backward view (Table~\ref{tab:ytraj-matching}). GLIMPSE achieves
the highest mean match and inlier counts, while GEN3C yields the lowest RGB MAE
over accepted correspondences. Matching details are provided in
Appendix~\ref{sec:experiment-protocol}.

Figure~\ref{fig:qualitative-overview} highlights cross-view inconsistencies:
UCM introduces a car and an additional street pole absent from the earlier
views, while SPMem changes the house's appearance. UCM and GEN3C also exhibit
spatial drift and local artifacts. GLIMPSE better preserves the scene layout
and appearance in these examples. Additional results are provided in
Appendix~\ref{sec:additional-ytraj-qualitative}.

\begin{figure}[t]
  \centering
  \includegraphics[width=\linewidth]{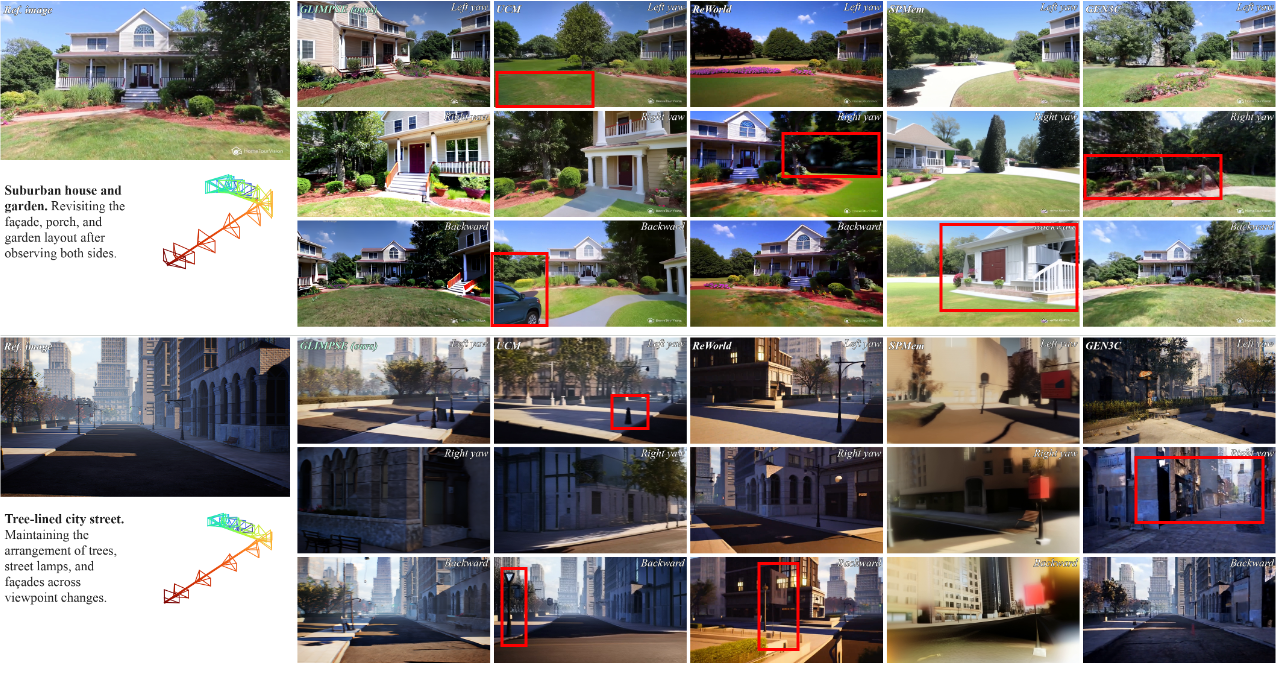}
  \caption{Y-traj. comparisons on house and street scenes. References and
  trajectories appear on the left, with left-yaw, right-yaw, and backward views
  shown top to bottom for each method. Red boxes highlight inconsistencies.}
  \label{fig:qualitative-overview}
\end{figure}

\subsection{Ablations}

\begin{figure}[t]
  \centering
  \includegraphics[width=\linewidth]{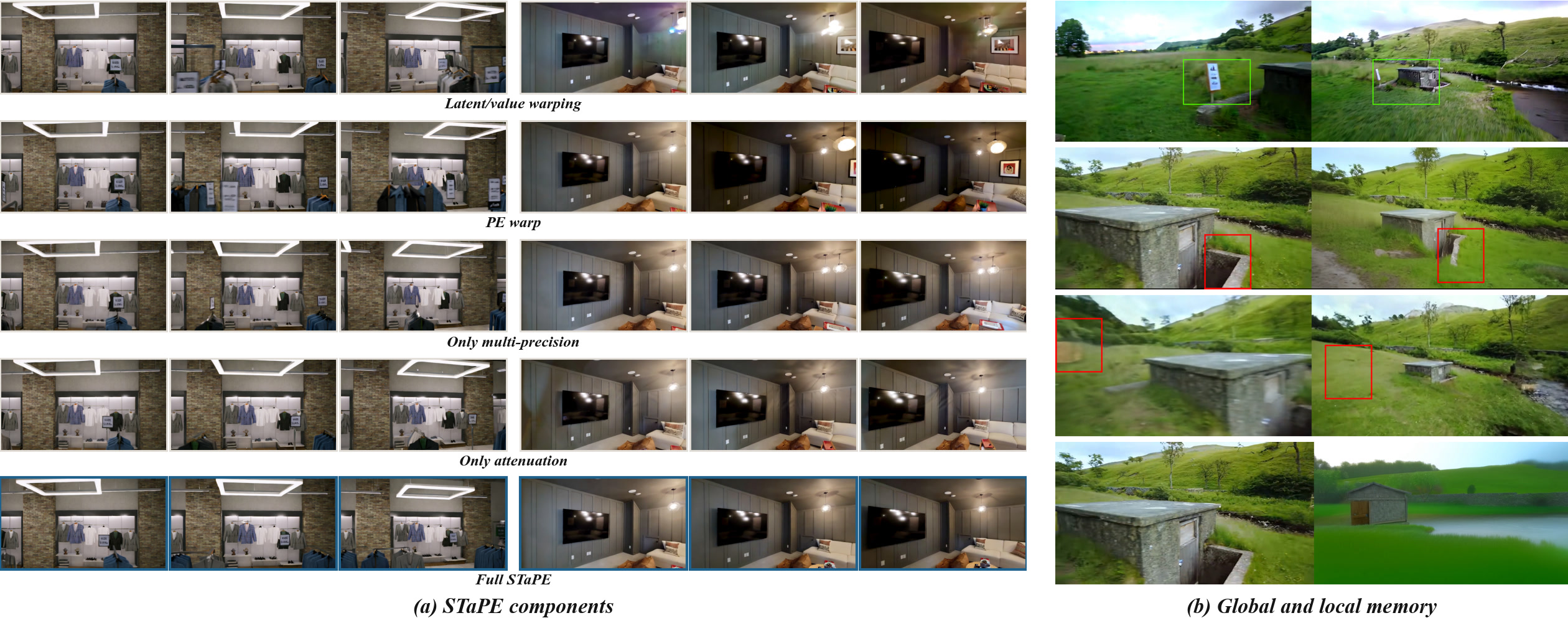}
  \caption{Qualitative ablations. \textbf{(a)} STaPE variants from
Table~\ref{tab:stape-ablation} on two held-out scenes. \textbf{(b)} Global--local
memory variants from Table~\ref{tab:memory-ablation} on a leave-and-return
trajectory. Green and red boxes highlight preserved content and
inconsistencies upon return, respectively.}
  \label{fig:stape-ablation}
\end{figure}

\begin{table}[!t]
\centering\scriptsize
\caption{STaPE component ablation. Multi-precision coordinates and spectral
attenuation are varied independently while footprint retrieval, local token
budget, and GRM remain fixed. The latent/value-warping control resamples
historical features onto the target grid and uses neither component.
Both ablation tables use the same separate evaluation set with several
trajectory types, including Y-traj. Rotation error is in degrees.}
\label{tab:stape-ablation}
\setlength{\tabcolsep}{1.5pt}
\begin{tabular}{lccrrrrrrrr}
\toprule
Configuration & \shortstack{Multi-\\precision} & \shortstack{Spectral\\attenuation}
& \shortstack{Mat. Pix.@5\\(K) $\uparrow$}
& \shortstack{Inliers@5\\(K) $\uparrow$}
& \shortstack{Inlier rate@5\\(\%) $\uparrow$}
& \shortstack{RGB\\MAE@5 $\downarrow$}
& \shortstack{Mean rot.\\($^\circ$) $\downarrow$}
& \shortstack{Trans.\\NRMSE $\downarrow$}
& CLIP-T $\uparrow$ & CLIP-F $\uparrow$ \\
\midrule
Latent/value warping & N/A & N/A & 6.410 & 5.530 & 86.271 & 0.121 & 1.27 & 0.270 & 27.08 & 96.11 \\
PE warp & Off & Off & 6.407 & 5.690 & 88.002 & 0.135 & 1.30 & 0.213 & 29.314 & 97.43 \\
Only multi-precision & On & Off & 6.567 & 5.707 & 87.656 & 0.122 & 1.19 & 0.210 & 29.618  & 97.71 \\
Only attenuation & Off & On & 6.917 & 6.097 & 88.145 & 0.120 & 1.25 & 0.212 & 29.780 & 98.57 \\
Full STaPE & On & On & 7.153 & 6.480 & 90.587 & 0.109 & 1.16 & 0.204 & 30.089 & 99.13 \\
\bottomrule
\end{tabular}
\end{table}


\begin{table}[!t]
\centering\scriptsize
\caption{Global recurrent (GRM) and local pointer (LPM) memory ablation.
Source, text, and camera conditions are retained. LPM uses full STaPE at a
fixed read budget, with the protocol of Table~\ref{tab:stape-ablation}.}
\label{tab:memory-ablation}
\setlength{\tabcolsep}{2pt}
\begin{tabular}{lccrrrrrrrr}
\toprule
Configuration & GRM & LPM
& \shortstack{Mat. Pix.@5\\(K) $\uparrow$}
& \shortstack{Inliers@5\\(K) $\uparrow$}
& \shortstack{Inlier rate@5\\(\%) $\uparrow$}
& \shortstack{RGB\\MAE@5 $\downarrow$}
& \shortstack{Mean rot.\\($^\circ$) $\downarrow$}
& \shortstack{Trans.\\NRMSE $\downarrow$}
& CLIP-T $\uparrow$ & CLIP-F $\uparrow$ \\
\midrule
No memory & Off & Off & 5.393 & 4.693 & 87.021 & 0.183 & 1.27 & 0.316 & 29.771 & 97.50 \\
Global only & On & Off & 5.743 & 5.200 & 90.540 & 0.152 & 1.20 & 0.304 & 29.939 & 98.49 \\
Local only & Off & On & 6.637 & 6.007 & 90.507 & 0.133 & 1.26 & 0.298 & 30.012 & 97.99 \\
Global + local & On & On & 7.153 & 6.480 & 90.587 & 0.109 & 1.16 & 0.204 & 30.089 & 99.13  \\
\bottomrule
\end{tabular}
\end{table}

Tables~\ref{tab:stape-ablation}--\ref{tab:memory-ablation} isolate STaPE and
the memory branches. Additional diagnostics support frequency-selective,
token-specific modulation over matched scalar suppression and permuted widths
(Tables~\ref{tab:stape-kv-diagnostic}--\ref{tab:stape-reliability-diagnostic}).
Under identical address perturbations, STaPE retains more geometric inliers
and lower RGB MAE than continuous PE transport
(Table~\ref{tab:stape-address-perturbation}).

\section{Conclusion and Limitations}
\label{sec:conclusion}
\label{sec:limitations}

We presented GLIMPSE, a global--local memory indexing framework with
multi-precision spectral encoding for explorable video world models.
Local pointer memory preserves view-specific historical
evidence, while recurrent global memory supplies complementary scene context.
At the core of this interface, STaPE addresses a challenge that persists even
when historical content is preserved: inaccurate geometry can assign that
content an overly precise positional address. By combining band-wise
multi-precision coordinates with geometry-conditioned spectral modulation,
STaPE improves memory reads without resampling historical appearance or
increasing the local token budget. Camera-control and Y-trajectory experiments
demonstrate improved cross-view correspondence and recovery of previously
observed content. Matched component ablations support the complementary roles
of the two addressing mechanisms and memory branches, while controlled address
perturbations show reduced sensitivity to projection errors. Together, these
results highlight that consistent exploration depends not only on retaining
relevant history, but also on how its geometric addresses enter attention.

However, the current system focuses on camera-conditioned exploration and does not
explicitly model object motion or use few-step distillation for real-time
inference. Extending memory updates to track changing object states and
distilling the generator to reduce denoising latency are natural next steps
toward more responsive exploration and interaction.

\bibliography{references}
\bibliographystyle{preprint}
\appendix
\section{Retrieval and Positional-Encoding Details}
\label{sec:retrieval-details}

\subsection{Spectral Regularization}
\label{sec:spectral-scope}
\paragraph{Operator settings.}
After projection and head reshaping, tensors have layout
$[B,L,N_{\mathrm{head}},D]$. Positional operators act independently on each
head's $D$ channels before scaled dot-product attention, without mixing heads
or token rows. The output transform $B_q^O$ precedes head concatenation and
output projection. Each token's $\sigma_h$ is shared across heads.
We use $\lambda_c=0.75$ and $\lambda_a=0.25$, with widths measured in
transported patch-grid units and depth differences computed on the aligned
historical depth grid. These coefficients give $\sigma_h\in[0,1]$, and
the operator characterization covers $\sigma\in[0,2]$.

\paragraph{Continuous-coordinate motivation.}
For fixed source and target cameras and fixed content, suppose only the continuous transported
coordinate is perturbed by $\epsilon\sim\mathcal N(0,\sigma^2I_2)$.
The Gaussian characteristic function gives
\begin{equation}
 \mathbb E[R_k(a+\epsilon_a)]=
 e^{-\sigma^2\omega_k^2/2}R_k(a),\qquad
 \mathbb E[B_h^{KV}(a+\epsilon)K^{\mathrm{pre}}]
 =D_\sigma B_h^{KV}(a)K^{\mathrm{pre}}.
 \label{eq:stape-expectation}
\end{equation}
The same conditional identity holds for $V^{\mathrm{pre}}$. It follows blockwise
because the camera block is fixed and the two spatial blocks are axial
Fourier pairs, as in integrated positional encoding~\citep{mipnerf}.

\paragraph{Measured address error.}
The width is intended to be larger where a transported address is less
reliable, which is testable without ground-truth geometry. For a held-out
sample of 60 clips we pair a history frame with a later frame at separations of
2, 4 and 8 latent steps and run GIM-DKM~\citep{gim,dkm} dense matching between the two real
RGB frames, retaining correspondences above 0.9 confidence. The matched
location of each patch centre is an estimate of where that surface actually
appears in the queried view, obtained without any estimated depth or pose, so
it is independent of the geometry that produced the transported coordinate.
Comparing it against the DA3 projection over 41{,}661 retained patch centres,
the median address error is 0.37 patch cells and $\sigma_h$ increases with it
(Spearman $\rho=0.329$, Pearson $r=0.383$, $p<10^{-150}$).

\paragraph{Reference settings.}
\begin{center}\small
\begin{tabular}{p{0.27\linewidth}p{0.65\linewidth}}
\hline
Candidate limits & Up to 16 source and 16 generated views per target, and
the all-source inference option removes only the source cap.\\
Footprint score & $4H_p\times4W_p$ grid, 16 samples per patch, and
source-priority selection exhausts positive gains.\\
Local read & $M=H_pW_p$ original tokens per target frame, and frame-local
adapter mask, distinct from native temporal attention.\\
Layer schedule & Interaction stride 1, and adapter width 3072, 24 heads,
128 channels per head.\\
Initialization & Pointer/global scalar gates and outer output projection
start at zero, and adapter input projection starts at identity.\\
\hline
\end{tabular}
\end{center}
The following sections specify source-priority retrieval, camera ordering,
token allocation, and positional encoding.

\paragraph{Three resolutions and two selections.}
Let the isomorphic patch grid be $H_p\times W_p$, after the shared DiT
patch embedding, with $M=H_pW_p$ tokens per frame. Source RGB pixels, these
patch cells, and the $4H_p\times4W_p$ footprint scoring grid are distinct.
Increasing the scoring resolution does not create additional attention tokens.
The footprint routine ranks candidate \emph{views}. A subsequent routine
chooses original \emph{tokens} from those views using center-cell coverage.
Consequently the area covered by all patches in selected views is not
necessarily the area supported by the final $M$ tokens.

\paragraph{Candidate bank and ordering.}
The bank contains decoded historical observations and their aligned DA3
depth/pose/intrinsics, with an immutable-source versus generated-history flag.
For each target frame, camera prefiltering uses
$s_i=\theta_i/(15^\circ)+\|\mathbf t_i-\mathbf t_q\|/0.25$,
where translation is in the scene-depth-normalized gauge.
The nearest view is first, and subsequent views are ordered by $s_i-\alpha_i$,
with stable observation order breaking ties. The default shortlist retains
at most 16 immutable-source and 16 generated views (16 total when no source
flag is supplied). The all-source inference option removes the former cap
but keeps the generated-view cap. Compared retrieval variants share this
setting and the upstream history sampling schedule, and
the attention budget remains $M$ independently of shortlist size.
The shortlist is per target frame, while projection can batch its union
across a target window.

View selection scores sampled geometric footprints, whereas positional
addressing uses the projected centers of the selected tokens, and the two stages
therefore consume different geometric primitives.

\paragraph{Projected footprint scoring.}
For each source patch, the routine unprojects its four grid corners using
one depth value shared by the corners and the grid-scaled source intrinsics,
then projects them into the target. A patch contributes only if all four
corners have positive target-camera depth ($>10^{-6}$).
Sixteen bilinear sample locations inside the projected quadrilateral are
rounded onto the $4H_p\times4W_p$ grid, and in-bounds samples mark occupied cells.
Their Boolean union defines the projected-occupancy mask $F_i$, counting
cross-view overlaps once. Table~\ref{tab:visibility-planned} compares this
retrieval strategy with center-hit and causal z-buffer controls.

\paragraph{Objective and implemented greedy policy.}
For a fixed bank, define
\begin{equation}
 C(S)=\frac{|\bigcup_{i\in S}F_i|}{16H_pW_p},\qquad
 \Delta(i\mid S)=\frac{|F_i\setminus\bigcup_{j\in S}F_j|}{16H_pW_p}.
 \label{eq:coverage-objective}
\end{equation}
This fixed-set union objective is monotone and submodular. The implemented
policy adds camera prefiltering, a seeded first view, source priority, and
separate token allocation on top of it.

The main-text retrieval-to-read procedure specifies the source-priority
branch. Equal marginal gains retain shortlist order, and accepting the seed
also updates $U$ with its footprint mask. Selection follows the source-priority
policy in the main text and exhausts positive coverage gains.

\paragraph{Token ownership, fillers, and PE.}
The budget allocator projects source-token centers and quantizes them by
$\lfloor u+0.5\rfloor,\lfloor v+0.5\rfloor$. In a collision, the first
token in selected-view/source-cell order receives coverage priority.
Remaining capacity is filled with unused original tokens, which may be
redundant, out of bounds, or behind the target camera.
Such fillers remain valid attention rows but do not count as target-visible
coverage. Invalid padding rows are excluded. A sign-preserving $10^{-6}$
denominator floor avoids division by zero, and out-of-frame positions are not
clipped into the target image. Each selected token retains source frame ID,
source cell ID, its continuous projected center, and reliability metadata.
Its PE uses the continuous projected center.

\paragraph{Tensor layout.}
For tensors $[B,L,N_h,D]$, the first $D/2$ channels are grouped into
four-channel projective vectors. In each following $D/4$ block, the two
halves form $D/8$ Fourier pairs, rather than adjacent even/odd channels.
The attenuation vector is repeated over these two halves and broadcast
over batch/head dimensions. It is applied after the projective/spatial
transform, only to pointer K/V spatial blocks. Equation~\ref{eq:stape-expectation}
follows for each sine and cosine component separately.

\subsection{Projection Convention and Layer Sequence}
\label{sec:projection-refresh}

Let $P_q$ and $P_h$ denote the lifted, intrinsics-aware world-to-camera matrices
of the target and historical token's source view in a shared scene gauge. We adopt the PRoPE head partition
$D/2+D/4+D/4$~\citep{prope}. The first block contains repeated four-dimensional
camera groups, and the remaining blocks contain horizontal and vertical rotary
pairs. Define
\begin{equation}
 \begin{split}
 B_q^Q(u,v)&=\operatorname{diag}\!\left(C(P_q^\top),R_x(u),R_y(v)\right),\\
 B_h^{KV}(u,v)&=\operatorname{diag}\!\left(C(P_h^{-1}),R_x(u),R_y(v)\right),\\
 B_q^O(u,v)&=\operatorname{diag}\!\left(C(P_q),R_x(u)^{-1},R_y(v)^{-1}\right).
 \end{split}
 \label{eq:stape-basis}
\end{equation}
Here $C(P)$ applies $P$ independently to each four-channel group.
Memory K/V retain their source-camera basis while their spatial coordinates
are projected into the target view. The camera blocks therefore contribute
$P_qP_h^{-1}$ to query--key interactions and value transport into the target basis.

\paragraph{Geometry and coordinate units.}
Forward depth differences in Equation~\ref{eq:stape-attenuation} are
zero-padded at the right and bottom boundaries. Absent explicit observation
times, history receives equally spaced times in sequence order.
$\operatorname{round}_{\mathrm{even}}$ in
Equation~\ref{eq:stape-footprint} is \texttt{torch.round}, not floor
quantization: $-1.5,-0.5,0.5,1.5$ map to $-2,0,0,2$.
Historical depth is bilinearly resized to the source patch grid with
\texttt{align\_corners=False}, and intrinsics are row-scaled to that grid before
unprojection, associating a resized depth sample with each patch center.
Source and target camera-to-world translations and depth must use a common
scene scale. Scene normalization translates the reference origin and divides
positions and depths by the same scene-depth scale, and it does not independently
rescale each camera. Target pixel coordinates $(u_{\rm pix},v_{\rm pix})$
become $u=u_{\rm pix}W_p/W_{\rm img}-1/2$ and
$v=v_{\rm pix}H_p/H_{\rm img}-1/2$.

\paragraph{Pointer-shadow update at layer $\ell$.}
Let $x^\ell$ denote the main-stream features and $p^\ell$ the packed
historical features. The shared patch embedding is applied once at input, and
subsequent layers operate on the updated hidden states.
For the pointer-shadow branch, the code executes the following sequence:
\begin{enumerate}
\item Use the current backbone block's normalization, Q/K/V projections,
and timestep modulation to process $p^\ell$ at timestep zero.
Pointer Q/K receive native DiT RoPE at their original source-grid coordinates, and
target-view transport is applied at the subsequent memory read.
\item Pointer queries attend to pointer tokens associated with the same
target frame, without reading noisy target features or other target-frame groups.
The same block's text attention and feed-forward weights produce
$\hat p^{\ell+1}$. A learned update gate yields
$p^{\ell+1}=p^\ell+\tanh(g_\ell)(\hat p^{\ell+1}-p^\ell)$.
\item Run the ordinary target backbone block on $x^\ell$.
Its native target-to-target temporal attention is retained. In this
shadow branch, this ordinary pass is distinct from appending pointer
tokens directly to one joint backbone sequence.
\item At configured adapter interaction layers, the updated target features
read the refreshed pointer features through the frame-local STaPE/PRoPE
operation in Equation~\ref{eq:stape-attention}.
CUT3R has a separate softmax, and its output is gated after attention, before
the combined conditioning update is projected back into the main stream.
\end{enumerate}
The reference pointer-shadow configuration uses \texttt{interact\_every=1},
so refresh and adapter read occur at every backbone layer. More generally,
stride $k$ selects zero-based blocks with $\ell\bmod k=k-1$.
Each adapter has width 3072, 24 heads of width 128, and its own parameters.
Its input projection is initialized to the identity, and Q/K/V and the inner
output projection copy the corresponding DiT attention weights at
initialization, without subsequent weight tying. Adapter LayerNorms remain
separate from the native block's normalization.

Both the pointer update parameter $g_\ell$ and the global-read parameter
$\beta_\ell$ are scalar, learned, and initialized to zero. If
$o_{L,\ell}$ is the local read after the inverse output-coordinate transform
and $o_{G,\ell}$ is the separately normalized global read, the target update is
\begin{equation}
 x_{\mathrm{target}}^{\ell+1}
 =\hat x_{\mathrm{target}}^{\ell+1}
 +B_\ell W_{O,\ell}
 \bigl(o_{L,\ell}+\tanh(\beta_\ell)o_{G,\ell}\bigr).
 \label{eq:adapter-residual}
\end{equation}
Here $\hat x^{\ell+1}$ is the ordinary backbone output and $B_\ell$ is
the zero-initialized outer output projection. There is no additional
scalar multiplying the local read in this \texttt{zero\_output} mode.
The source/TiER prefix is not modified by this adapter residual.
Each token's $\sigma_h$ is broadcast over attention heads.

\section{Training and Evaluation Details}
\label{sec:training-details}

\subsection{Noise Schedule and Prediction Convention}
\label{sec:flow-schedule}
The dense reference trainer samples an index uniformly from $N=1000$
scheduler nodes. Before numerical timestep casting, node $j$ is
$u_j=1-j/N$, $s_j=\rho u_j/[1+(\rho-1)u_j]$, and $t_j=1000s_j$,
for $j=0,\ldots,N-1$. The wrapper's reference shift is $\rho=5$.
The scheduler weights in Equation~\ref{eq:full-flow-matching} are
\begin{equation}
 y_j=\exp\!\left[-2\left(\frac{t_j-N/2}{N}\right)^2\right],\qquad
 w_j=\frac{N(y_j-\min_i y_i)}{\sum_i(y_i-\min_r y_r)}.
 \label{eq:flow-weights}
\end{equation}
The squared latent-space velocity prediction error is averaged over latent elements before applying
this weight. The ordinary training path casts the model timestep to BF16, and
noise construction and loss weighting use the nearest stored scheduler
node to that supplied timestep. This numerical lookup does not redefine
the velocity as a derivative with respect to the timestep index.
The displayed rollouts use 50 Euler steps with the shifted schedule and
a final update to zero noise, and no stochastic SDE term is added.

\paragraph{Future donors.}
The future augmentation is training-only. With target start $s$ and
length $T=20$ on the latent-aligned observation clock, the trainer checks
every donor index $j>s+T+g$. We use $g=40$,
donor sampling stride 15, at most 16 donor observations, and attempt
probability 0.70. These are observation-index units, not seconds, and frame
rate and temporal subsampling determine their physical duration.
The reader selects available donors beyond this guard and supplies their
indices explicitly, and the trainer does not assume a uniform continuous-time
distribution. Donors are sampled only when eligible observations are available.
RGB, depth, pose, and intrinsics retain the same true donor index.
Only the timestamps used to compute memory temporal lag are relabeled, and target RGB/cameras are unchanged.
No future donor is available in causal rollout inference.

\paragraph{Revisit and fast-motion augmentation.}
Synthetic revisit perturbs multiple selected memory views using a shared
camera-local delta, leaving target RGB and target camera commands unchanged.
Revisitation is attempted with probability 0.50, with perturbations up to
$45^\circ$ yaw, 0.20 scene-depth units laterally, and 0.12 units forward.
A joint-bank option permits future
donors to receive revisit perturbations as well, and these two augmentations
are not inherently mutually exclusive. Fast-motion augmentation uses
temporally subsampled targets with synchronously indexed RGB and geometry,
not a change of playback FPS alone.

Future donors provide additional visual support during training, and inference
uses causal observations. Training/test separation is defined at source-video
level and includes derived clips, donor pools, and geometry caches.

Table~\ref{tab:augmentation-ablation} isolates each augmentation by
retraining the complete system with it removed.

\begin{table}[t]
\centering\scriptsize
\caption{Training-augmentation ablation. Each row retrains the complete system
with the indicated augmentations and is evaluated on the same 100-case set
under the same causal inference protocol. The causal-only variant isolates
the contribution of training-time future-as-memory. Column definitions follow
Table~\ref{tab:stape-ablation}.}
\label{tab:augmentation-ablation}
\setlength{\tabcolsep}{2pt}
\resizebox{\linewidth}{!}{%
\begin{tabular}{lcccrrrrrrr}
\toprule
Configuration
& \shortstack{Future-as-\\memory}
& \shortstack{Synthetic\\revisit}
& \shortstack{Fast\\motion}
& \shortstack{Mat. Pix.@5\\(K) $\uparrow$}
& \shortstack{Inliers@5\\(K) $\uparrow$}
& \shortstack{RGB\\MAE@5 $\downarrow$}
& \shortstack{Mean rot.\\($^\circ$) $\downarrow$}
& \shortstack{Trans.\\NRMSE $\downarrow$}
& CLIP-T $\uparrow$ & CLIP-F $\uparrow$ \\
\midrule
All augmentations & On & On & On & 7.153 & 6.480  & 0.109 & 1.16 & 0.204 & 30.089 & 99.13 \\
No future-as-memory (causal only) & Off & On & On  & 6.851 & 5.918 & 0.134 & 1.53 & 0.288 & 28.172 & 97.12 \\
No synthetic revisit & On & Off & On  & 6.929 & 6.020 & 0.147 & 1.33 & 0.273 & 28.090 & 97.03 \\
No fast motion & On & On & Off  & 7.011 & 6.220 & 0.119 & 1.78 & 0.312 & 29.100 & 96.90 \\
No augmentation & Off & Off & Off & 6.629 & 5.644 &0.162 & 1.90 & 0.338 & 27.936 & 96.02 \\
\bottomrule
\end{tabular}}%
\end{table}

\subsection{Experimental Settings and Comparison Protocols}
\label{sec:experiment-protocol}

\paragraph{Training configuration.}
We train the backbone of Section~\ref{sec:stape}
at increasing resolution, from $256\times448$ to $352\times640$, with a
20-latent-frame target window.
Training uses a mixture of SpatialVID, RealCam-Vid, ReCamMaster, and
Context-as-Memory. Cached DA3~\citep{da3} pose, intrinsics, and depth
provide geometry, and frozen CUT3R with MeMix~\citep{cut3r,memix} provides
recurrent context. Internal controls hold initialization, data, optimization
budget, geometry, and augmentations fixed except for the factor under study.
We train for 40,000 optimization steps using AdamW with learning rate
$10^{-6}$ for the backbone and memory-injection modules, weight decay
$10^{-4}$, $\beta=(0.9,0.999)$, and $\epsilon=10^{-8}$.
Training uses eight GPUs with one sample per GPU and four gradient
accumulation steps, giving an effective batch size of 32.
The DiT backbone and memory-injection modules are optimized jointly, while
the VAE, text encoder, and recurrent geometry encoder remain frozen.

\paragraph{Tasks and inference settings.}
Camera replay follows trajectories estimated from the corresponding source
videos. Y-traj. observes the left and right sides before moving backward
at the initial heading, producing 240 new RGB frames plus the source frame.
Internal paired comparisons share the observed prefix, prompt, requested
cameras, seed list, resolution, sampler, and horizon. Inference donors consist
of observed source frames and completed generated windows. Generated endpoints
are retained as generated. Source-video-disjoint splits separate derived
clips, donor pools, and geometry caches.

\paragraph{STaPE controls.}
Table~\ref{tab:stape-ablation} holds the observation pool, cameras, footprint
retrieval, global branch, and local read budget fixed. Latent/value warping
resamples historical features onto the target grid. PE warp uses continuous
coordinates in every spatial band and disables spectral attenuation.
The two intermediate controls enable multi-precision coordinates or spectral
attenuation separately, and the full configuration enables both. Attenuation
modulates both memory keys and values.

\paragraph{Memory controls.}
Table~\ref{tab:memory-ablation} independently enables GRM and LPM. GRM supplies
frozen chronological recurrent context, and LPM supplies geometry-indexed patches
with full STaPE. Source, text, and camera conditioning are retained in all
four configurations. The local read budget is fixed whenever LPM is enabled.

\paragraph{Metrics and aggregation.}
The camera-control table reports mean and endpoint rotation errors,
translation NRMSE, translation-direction error, CLIP-T, CLIP-F, FID, and FVD.
DA3-derived pose scores quantify agreement with the requested trajectory
in the estimated scene coordinate system. WorldScore component scaling and composite weighting are given in the
table captions.
For per-case metrics, frame or frame-pair measurements are averaged within
each case, followed by an equal-weight average over cases. Inlier fractions
are computed per pair before this aggregation. For revisit correspondences, each video pairs frames 76--80 and
156--160 with frames 236--240, using zero-based one-to-one indices, and GIM-DKM~\citep{gim,dkm}
samples 15,000 correspondences per pair after a common $240\times416$ resize
and padding, and matches above 0.1 confidence are counted (Mat. Pix.).
Geometric inliers are estimated with a fundamental matrix using OpenCV
USAC\_MAGSAC, a 1-pixel threshold in the unpadded $240\times416$ coordinates,
confidence 0.999999, and at most 10,000 iterations. Inliers are counted only
among matches above 0.1 confidence, and RGB MAE is evaluated at these accepted
correspondences using bilinear sampling of RGB values in $[0,1]$.
FID and FVD are computed over the evaluation set as
distribution-level metrics. Motion magnitude measures the median displacement
of estimated optical flow rather than camera accuracy, and flow consistency
measures forward--backward optical-flow agreement. On the Y-traj. set,
all six methods use the full shared case set and the same evaluation protocol.
Appendix~\ref{sec:validation-records} defines
the counting units and aggregation details.

\paragraph{External comparisons.}
The experiments compare GLIMPSE with the state-of-the-art systems that have
public implementations, namely UCM, SPMem, GEN3C, VMem, and ReWorld. We follow
the inference recipes provided in their public repositories, adapting input
and camera interfaces to the evaluation tasks. The system-level tables compare complete
generation pipelines, and the internal ablations isolate changes within GLIMPSE's
shared backbone and training recipe.

\paragraph{VMem configuration.}
VMem uses a single source image, native
$576\times576$ generation, 50 sampling steps, guidance 2.0 (minimum 1.2),
four context and four target frames per model call, and 400 surfel iterations.
The source raster is resized with matching intrinsics scaling, and camera axes
are converted at the adapter boundary. WorldScore uses frames 0 and 80 as
anchors and the original source image as the style reference.

\subsection{Evidence Records and Aggregation}
\label{sec:validation-records}

\paragraph{Diagnostic scope.}
Projected footprint occupancy measures image-plane support. Target-depth
agreement additionally tests positive depth, image bounds, and agreement
with a reference surface using reference target depth for evaluation.
The fixed budget bounds local attention read length, and
geometry estimation, retrieval, transfer, denoising, and storage account
for the remaining system costs.

\ifdefined\graspinternalfigures
\paragraph{Existing qualitative record.}
The displayed panel contains two distinct source videos and two selected
completed rollouts, one per source, at
$352\times640$, with an 81-frame observed source and seeds 211158827 and
1610094911. It is a selected qualitative panel, not a randomly sampled
evaluation set, and its two completed cases do not establish a completion rate.
The saved videos contain 77 decoded frames per generated window, and the panels
use source frame 80 and window-local frame 76. Each example includes three
nominal yaw windows and one additional corrective return window. This
exported-frame count is specific to these recordings, not evidence for the
nominal 240-new-frame quantitative protocol. Original decoded images,
video hashes, case paths, and the unmodified correspondence diagnostic
are retained in the accompanying diagnostic manifest.
These two examples document the implementation and its residual drift, and
they are not the evaluation population for the quantitative tables.
\fi

\subsection{STaPE Mechanism Validation}
\label{sec:stape-mechanism-validation}

\paragraph{Key and value modulation.}
We compare modulation of keys, values, and both, retaining footprint retrieval,
global memory, and the local read budget. Multi-precision coordinates remain enabled.
To separate frequency-selective modulation from scalar suppression, the scalar control
replaces the frequency-dependent multipliers by their per-token mean on the
same spatial components of K and V, providing a magnitude-control baseline.

\begin{table}[htbp]
\centering
\begin{minipage}[t]{0.52\linewidth}
\vspace{0pt}
\caption{Inference-time attenuation-path comparison. None disables attenuation.}
\label{tab:stape-kv-diagnostic}
\end{minipage}\hfill
\begin{minipage}[t]{0.46\linewidth}
\vspace{0pt}
\caption{Reliability-signal comparison. Coordinate precision and joint K/V
modulation are fixed. Depth + temporal lag uses full STaPE.}
\label{tab:stape-reliability-diagnostic}
\end{minipage}
\par\smallskip
\begin{minipage}[t]{0.52\linewidth}
\vspace{0pt}\centering\scriptsize
\setlength{\tabcolsep}{2.5pt}
\begin{tabular}{lccc}
\toprule
Modulation & \shortstack{Mat. Pix.@5\\(K)$\uparrow$} & \shortstack{Inliers@5\\(K)$\uparrow$} & \shortstack{RGB MAE\\@5$\downarrow$} \\
\midrule
None & 6.567 & 5.707 & 0.122 \\
K only & 6.650 & 5.985 & 0.114 \\
V only & 6.501 & 6.199 & 0.125 \\
K + V (STaPE) & 7.153 & 6.480 & 0.109 \\
Matched scalar on K + V & 6.864 & 6.272 & 0.119 \\
\bottomrule
\end{tabular}
\end{minipage}\hfill
\begin{minipage}[t]{0.46\linewidth}
\vspace{0pt}\centering\scriptsize
\setlength{\tabcolsep}{3pt}
\begin{tabular}{lcc}
\toprule
Width signal & \shortstack{Mat. Pix.@5\\(K)$\uparrow$} & \shortstack{RGB MAE\\@5$\downarrow$} \\
\midrule
Depth only & 6.799 & 0.126 \\
Temporal lag only & 6.723 & 0.119 \\
Depth + temporal lag & 7.153 & 0.109 \\
\shortstack[l]{Permuted depth +\\temporal lag} & 6.692 & 0.194 \\
\bottomrule
\end{tabular}
\end{minipage}
\end{table}

Joint K/V modulation achieves the highest correspondence counts. Compared
with the matched scalar control, STaPE increases geometric inliers by 3.32\%
and reduces RGB MAE by 8.40\%, supporting frequency-selective attenuation.

\paragraph{Depth and temporal-lag signals.}
The reliability diagnostic compares depth-only, temporal-lag-only, and combined widths,
and a control that permutes the combined widths among retrieved tokens within
each query, preserving their distribution. Coefficients retain their default
values, $\lambda_c=0.75$ and $\lambda_a=0.25$, when the other term is removed.
The combined row uses full STaPE.

Combining depth and temporal lag improves both metrics over either signal alone.
Relative to permuted widths, the combined signal increases matched pixels by
6.89\% and reduces RGB error by 43.81\%, supporting token-specific modulation.

\paragraph{Sensitivity to address perturbations.}
We compare continuous PE transport and full STaPE using identical cases,
generation seeds, and retrieved tokens. Independent zero-mean Gaussian offsets with standard deviation $s$
are added to the two projected spatial coordinates before positional encoding
and, for STaPE, before quantization. Both variants receive identical offsets,
held fixed throughout denoising. Retrieval, content, and STaPE's depth- and temporal-lag-conditioned
widths remain unchanged, isolating sensitivity to address errors.

\begin{table}[htbp]
\centering\small
\caption{Address perturbations with standard deviation $s$ in patch-grid units.}
\label{tab:stape-address-perturbation}
\setlength{\tabcolsep}{5pt}
\begin{tabular}{clccc}
\toprule
$s$ & Addressing & Mat. Pix.@5 (K)$\uparrow$ & Inliers@5 (K)$\uparrow$ & RGB MAE@5$\downarrow$ \\
\midrule
0 & PE warp & 6.407 & 5.690 & 0.135 \\
  & STaPE & 7.153 & 6.480 & 0.109 \\
\midrule
0.25 & PE warp & 6.043 & 5.205 & 0.159 \\
     & STaPE & 7.100 & 6.311 & 0.120 \\
0.50 & PE warp & 6.019 & 4.897 & 0.183 \\
     & STaPE & 6.894 & 6.242 & 0.139 \\
1.00 & PE warp & 5.229 & 4.288 & 0.216 \\
     & STaPE & 6.245 & 5.776 & 0.145 \\
\bottomrule
\end{tabular}
\end{table}

STaPE retains more matched pixels and geometric inliers, with lower RGB MAE,
at every tested nonzero perturbation level. At $s=1$, it yields 19.4\% more
matched pixels and 34.7\% more inliers than PE warp, while reducing RGB MAE
by 32.9\%, supporting robust memory reads under perturbed positional addresses.

\subsection{Retrieval Support and Exploration Cost}
\label{sec:retrieval-scaling}

\paragraph{Retrieval support.}
Tables~\ref{tab:retrieval-support-archive} and~\ref{tab:retrieval-cost-archive}
summarize six rollouts from five sources, with 220 tokens per target frame.
Return denotes an additional corrective window. Unique cells count distinct
in-bounds cells relative to the target grid, and repeats count excess tokens in
occupied cells relative to the read budget. Neither measures visibility.

\begin{table}[htbp]
\centering
\begin{minipage}[t]{0.52\linewidth}
\vspace{0pt}
\caption{Retrieval support (\%). Sources are equally weighted, and
brackets give 95\% source-cluster bootstrap intervals.
Source share denotes tokens from original observations.}
\label{tab:retrieval-support-archive}
\end{minipage}\hfill
\begin{minipage}[t]{0.46\linewidth}
\vspace{0pt}
\caption{Median component times over six rollouts. DA3 inputs include source
and generated observations. Other pipeline stages are excluded.}
\label{tab:retrieval-cost-archive}
\end{minipage}
\par\smallskip
\begin{minipage}[t]{0.52\linewidth}
\vspace{0pt}\centering\scriptsize
\setlength{\tabcolsep}{2pt}
\begin{tabular}{lrrrr}
\toprule
Window & \shortstack{In bounds\\{[95\% CI]}} & \shortstack{Unique\\cells} & Repeats & \shortstack{Source\\share} \\
\midrule
1 & 61.2 [45.9, 76.4] & 59.9 & 1.2 & 100.0 \\
2 & 68.3 [62.1, 73.2] & 60.0 & 8.3 & 81.4 \\
3 & 75.0 [64.2, 86.6] & 72.8 & 2.1 & 79.1 \\
Return & 95.2 [90.1, 98.6] & 94.6 & 0.6 & 87.4 \\
\bottomrule
\end{tabular}
\end{minipage}\hfill
\begin{minipage}[t]{0.46\linewidth}
\vspace{0pt}\centering\scriptsize
\setlength{\tabcolsep}{4pt}
\begin{tabular}{lrrr}
\toprule
Window & \shortstack{DA3\\observations} & \shortstack{DA3\\forward (s)} & \shortstack{Pointer\\cache (s)} \\
\midrule
1 & 41 & 1.68 & 3.61 \\
2 & 61 & 2.72 & 2.37 \\
3 & 81 & 4.14 & 3.92 \\
Return & 101 & 5.47 & 4.23 \\
\bottomrule
\end{tabular}
\end{minipage}
\end{table}

For a single 241-frame revisit rollout, the third window contains 20 latent
queries with 220 retrieved tokens each. In-bounds tokens account for 95.41\%
of the read budget, distinct in-bounds cells cover 94.27\% of the target grid,
and repeated in-bounds projections account for 1.14\% of the read budget.

\paragraph{Visibility diagnostic.}
The comparison fixes donors, cameras, checkpoint, seeds, and token
budget. Valid support counts positive-depth, in-bounds tokens relative to
the read budget. Depth consistency is the fraction of reference-valid
projections with $|z_{\mathrm{proj}}-z_{\mathrm{ref}}|/z_{\mathrm{ref}}\leq0.1$.
The causal z-buffer uses observed history for selection. Reference target
depth is used only to evaluate depth consistency.

\paragraph{Horizon profiling.}
Table~\ref{tab:horizon-cost-planned} reports cumulative denoising and DA3
forward times for 81-, 241-, and 721-frame rollouts at $352\times640$,
50 denoising steps per window, and CFG 5.

\begin{table}[htbp]
\centering
\begin{minipage}[t]{0.55\linewidth}
\vspace{0pt}
\caption{Retrieval comparison.
Support columns are percentages, and @5 follows Table~\ref{tab:ytraj-matching}.}
\label{tab:visibility-planned}
\end{minipage}\hfill
\begin{minipage}[t]{0.43\linewidth}
\vspace{0pt}
\caption{Cumulative component times (seconds).
Profiling includes diagnostic recording. DA3 forward times exclude model loading.}
\label{tab:horizon-cost-planned}
\end{minipage}
\par\smallskip
\begin{minipage}[t]{0.55\linewidth}
\vspace{0pt}\centering\scriptsize
\setlength{\tabcolsep}{2.5pt}
\begin{tabular}{lccc}
\toprule
Selector & \shortstack{Valid\\(\%)} & \shortstack{Depth\\consistent (\%)} & \shortstack{Inliers@5\\(K)} \\
\midrule
Footprint & 92.09 & 37.00 & 6.923 \\
Center-hit & 90.18 & 31.59 & 6.030 \\
Causal z-buffer & 89.75 & 36.21 & 6.852 \\
\bottomrule
\end{tabular}
\end{minipage}\hfill
\begin{minipage}[t]{0.43\linewidth}
\vspace{0pt}\centering\scriptsize
\setlength{\tabcolsep}{2pt}
\begin{tabular}{rcc}
\toprule
Frames & Denoising (s) & DA3 (s) \\
\midrule
81 & 174.26 & 1.33 \\
241 & 533.40 & 7.52 \\
721 & 1580.41  & 55.92 \\
\bottomrule
\end{tabular}
\end{minipage}
\end{table}

\section{Reproducibility Statement}
\label{sec:reproducibility}

We have submitted the core code and visualization videos as supplementary
materials. The full code and model weights will be publicly released upon
acceptance.

\paragraph{AI use statement.}
We used AI tools to polish the manuscript's language and assist in searching
for related literature.

\clearpage
\section{Additional Qualitative Results}
\label{sec:additional-ytraj-qualitative}
\label{sec:pointcloud-comparison}

\begin{figure}[!ht]
  \centering
  \includegraphics[width=\linewidth,height=0.35\textheight,keepaspectratio]{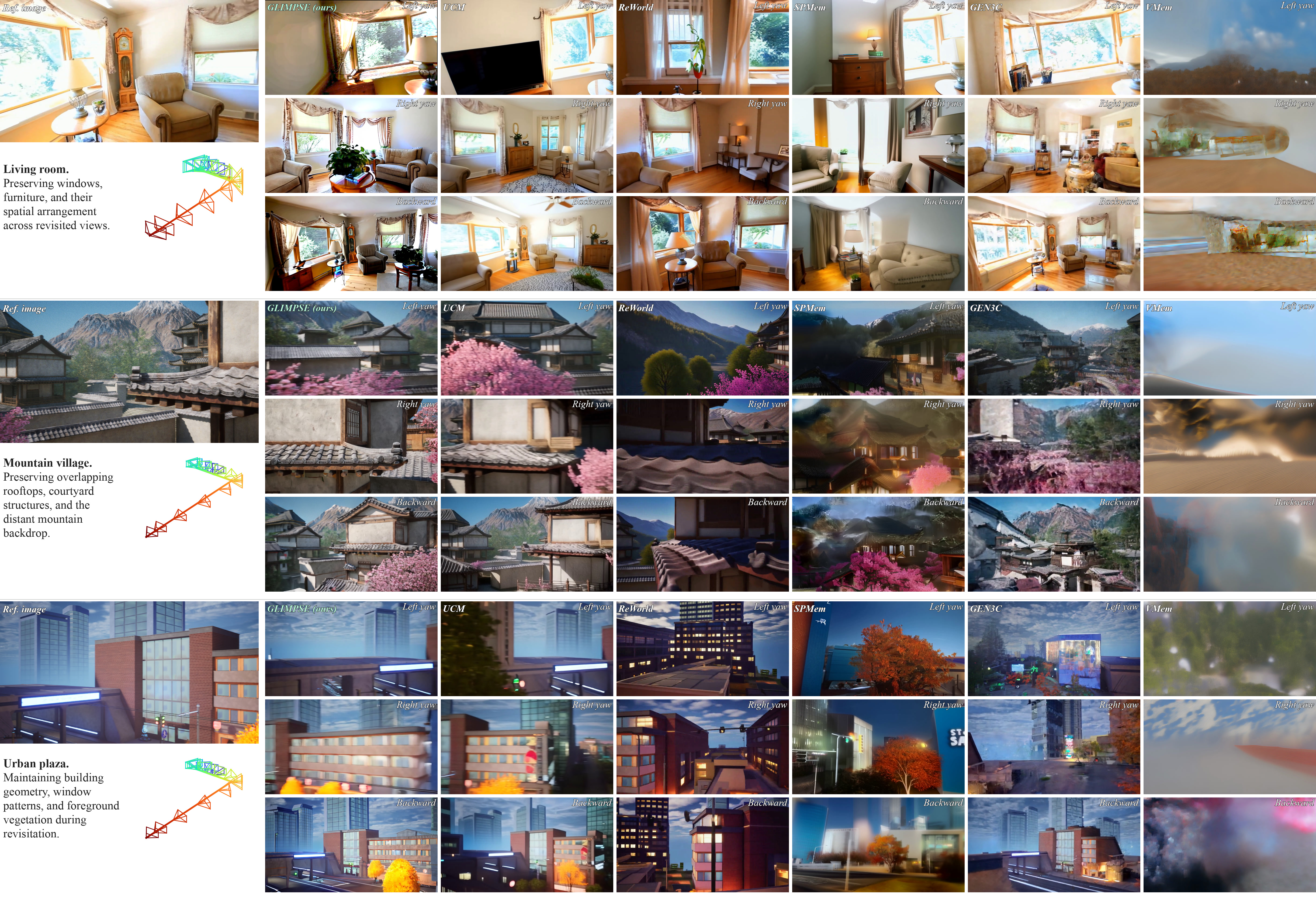}
  \caption{Additional Y-traj. comparisons in a living room, mountain village,
  and urban plaza, following the layout of Figure~\ref{fig:qualitative-overview}.
  Each scene shows left-yaw, right-yaw, and backward views across six methods.}
  \label{fig:additional-ytraj-qualitative}
\end{figure}

\begin{figure}[!ht]
  \centering
  \includegraphics[width=0.9\linewidth]{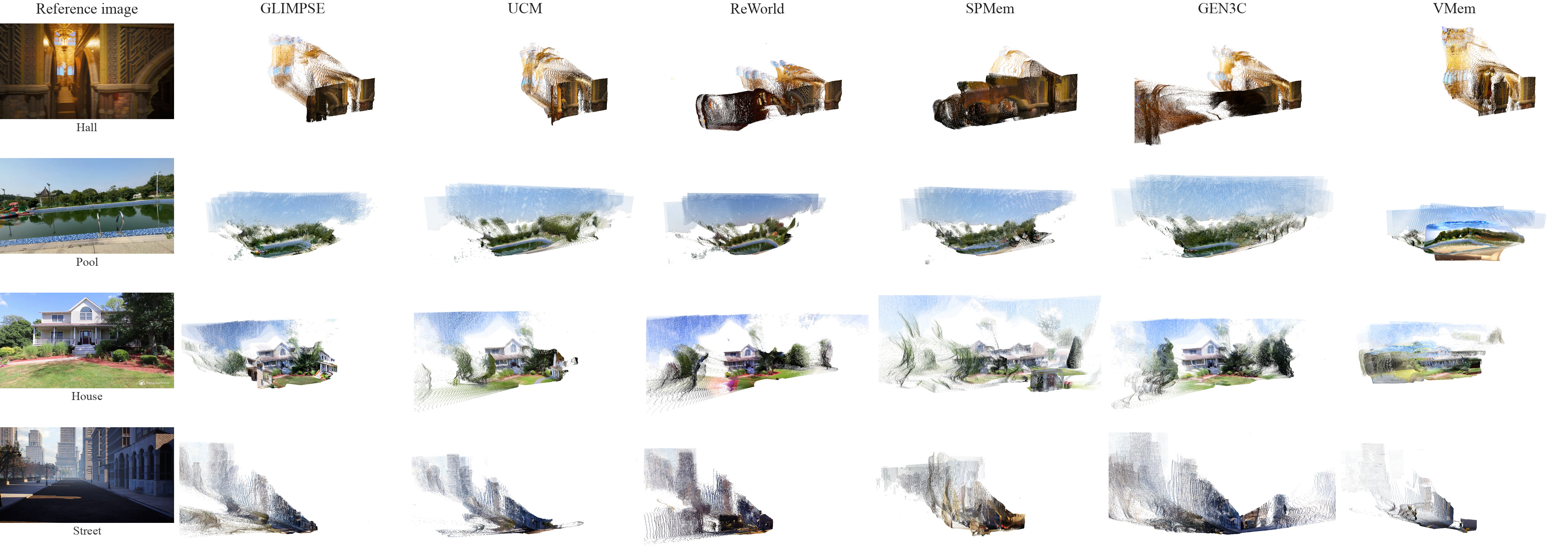}
  \caption{DA3 reconstructions of generated videos, colored by generated RGB.
  Methods share reconstruction settings, source-depth alignment, and a common
  rendering viewpoint and display scale per scene. We sample frames 8--80 at
  stride 8 for the hall and pool, and 16--224 at stride 16 for the house and
  street. The shared source anchors reconstruction but is not displayed.}
  \label{fig:pointcloud-comparison}
\end{figure}

\end{document}